\documentclass[11pt,a4paper]{article}
\usepackage[hyperref]{acl2018}
\usepackage{times}
\usepackage{latexsym}
\usepackage{graphicx}
\usepackage{amsmath}
\usepackage{amssymb}
\usepackage{multirow}
\usepackage{makecell}
\usepackage{url}

\usepackage{verbatim}
\usepackage{url}
\usepackage{array}
\usepackage{balance}
\usepackage{color}
\usepackage{soul}
\usepackage{hyperref}
\newcolumntype{L}[1]{>{\raggedright\let\newline\\\arraybackslash\hspace{0pt}}m{#1}}
\setul{2pt}{2pt}

\usepackage{pifont}
\newcommand{\xmark}{\ding{55}}

\aclfinalcopy 

\newcommand*{\affaddr}[1]{#1} 
\newcommand*{\affmark}[1][*]{\textsuperscript{#1}}
\newcommand*{\email}[1]{\texttt{#1}}

\title{Transformation Networks for Target-Oriented Sentiment Classification\thanks{\ \ The work was done when Xin Li was an intern at Tencent AI Lab. This project is substantially supported by a grant from the Research Grant Council of the Hong Kong Special Administrative Region, China (Project Code: 14203414).}}

\author{%
Xin Li\affmark[1], Lidong Bing\affmark[2], Wai Lam\affmark[1] and
Bei Shi\affmark[1]\\
\affaddr{\affmark[1]Department of Systems Engineering and Engineering Management\\
The Chinese University of Hong Kong, Hong Kong}\\
\affaddr{\affmark[2]Tencent AI Lab, Shenzhen, China}\\
\email{\{lixin,wlam,bshi\}@se.cuhk.edu.hk}\\
\email{lyndonbing@tencent.com}\\
}

\date{}

\begin{document}
\maketitle
\begin{abstract}
 Target-oriented sentiment classification aims at classifying sentiment polarities over individual opinion targets in a sentence. RNN with attention seems a good fit for the characteristics of this task, and indeed it achieves the state-of-the-art performance. After re-examining the drawbacks of attention mechanism and the obstacles that block CNN to perform well in this classification task, we propose a new model to overcome these issues. Instead of attention, our model employs a CNN layer to extract salient features from the transformed word representations originated from a bi-directional RNN layer. Between the two layers, we propose a component to  generate target-specific representations of words in the sentence, meanwhile incorporate a mechanism for preserving the original contextual information from the RNN layer. Experiments show that our model achieves a new state-of-the-art performance on a few benchmarks.\footnote{Our code is open-source and available at \url{https://github.com/lixin4ever/TNet}} 
 
\end{abstract}

\vspace{2mm}

\section{Introduction}
\label{sec:intro}
Target-oriented (also mentioned as ``target-level'' or ``aspect-level'' in some works) sentiment classification aims to determine sentiment polarities over ``opinion targets'' that explicitly appear in the sentences~\cite{liu2012sentiment}. For example, in the sentence \textit{``I am pleased with the fast \textbf{log on}, and the long \textbf{battery life}''}, the user mentions two targets ``\textit{\textbf{log on}}'' and ``\textit{\textbf{better life}}'', and expresses positive sentiments over them. 
The task is usually formulated as predicting a sentiment category for a (target, sentence) pair. 

Recurrent Neural Networks (RNNs) with attention mechanism, firstly proposed in machine translation~\cite{bahdanau2014neural}, is the most commonly-used technique for this task. For example, \citet{wang-EtAl:2016:EMNLP20163,tang-qin-liu:2016:EMNLP2016,yang2017attention,liu-zhang:2017:EACLshort,ma2017interactive} and \citet{chen-EtAl:2017:EMNLP20171} employ attention to measure the semantic relatedness between each context word and the target, and then use the induced attention scores to aggregate contextual features for prediction. 
In these works, the attention weight based combination of word-level features for classification may introduce noise and downgrade the prediction accuracy. For example, in ``\textit{This \textbf{dish} is my favorite and I always get it and never get tired of it.}'', these approaches tend to involve irrelevant words such as ``never'' and ``tired'' when they highlight the opinion modifier ``favorite''. To some extent, this drawback is rooted in the attention mechanism, as also observed in machine translation~\cite{luong-pham-manning:2015:EMNLP} and image captioning~\cite{xu2015show}.

Another observation is that the sentiment of a target is usually determined by key phrases such as ``is my favorite''. By this token, Convolutional Neural Networks (CNNs)---whose capability for extracting the informative n-gram features (also called ``active local features'') as sentence representations has been verified in~\cite{kim:2014:EMNLP2014,johnson2015semi}--- should be a suitable model for this classification problem. 
However, CNN likely fails in cases where a sentence expresses different sentiments over multiple targets, such as \textit{``great \textbf{food} but the \textbf{service} was dreadful!''}. One reason is that CNN cannot fully explore the target information as done by RNN-based methods \cite{tang-EtAl:2016:COLING3}.\footnote{One method could be concatenating the target representation with each word representation, but the effect as shown in \cite{wang-EtAl:2016:EMNLP20163} is limited.}  Moreover, it is hard for vanilla CNN to differentiate opinion words of multiple targets. Precisely, multiple active local features holding different sentiments (e.g., ``great food'' and ``service was dreadful'') may be captured for a single target, thus it will hinder the prediction. 

We propose a new architecture, named Target-Specific \textbf{T}ransformation \textbf{Net}works (TNet), to solve the above issues in the task of target sentiment classification. TNet firstly encodes the context information into word embeddings and generates the contextualized word representations with LSTMs. To integrate the target information into the word representations, TNet introduces a novel Target-Specific Transformation (TST) component for generating the target-specific word representations. Contrary to the previous attention-based approaches which apply the same target representation to determine the attention scores of individual context words, TST firstly generates different representations of the target conditioned on individual context words, then it consolidates each context word with its tailor-made target representation to obtain the transformed word representation.
Considering the context word \textit{``long''} and the target ``\textit{battery life}'' in the above example, TST firstly measures the associations between ``long'' and individual target words. Then it uses the association scores to generate the target representation conditioned on ``long''. After that, TST transforms the representation of ``long'' into its target-specific version with the new target representation. Note that ``long'' could also indicate a negative sentiment (say for ``startup time''), and the above TST is able to differentiate them.


As the context information carried by the representations from the LSTM layer will be lost after the non-linear TST, we design a context-preserving mechanism to contextualize the generated target-specific word representations. Such mechanism also allows deep transformation structure to learn abstract features\footnote{Abstract features usually refer to the features ultimately useful for the task~\cite{bengio2013representation,lecun2015deep}.}. 
To help the CNN feature extractor locate sentiment indicators more accurately, we adopt a proximity strategy to scale the input of convolutional layer with positional relevance between a word and the target.

In summary, our contributions are as follows: 

\indent $\bullet$   TNet adapts CNN to handle target-level sentiment classification, and its performance dominates the state-of-the-art models on benchmark datasets.\\
\indent $\bullet$   A novel Target-Specific Transformation component is proposed to better integrate target information into the word representations.
\\
\indent $\bullet$  A context-preserving mechanism is designed to forward the context information into a deep transformation architecture, thus, the model can learn more abstract contextualized word features from deeper networks.

\section{Model Description}
Given a target-sentence pair $(\mathrm{\bf w}^ \tau,\mathrm{\bf w})$, where $\mathrm{\bf w}^\tau = \{w^{\tau}_1,  w^{\tau}_2, ..., w^{\tau}_m\}$ is a sub-sequence of $\mathrm{\bf w}= \{w_1,  w_2, ..., w_n\}$, and the corresponding word embeddings $\mathrm{\bf x}^{\tau}=\{x^{\tau}_1, x^{\tau}_2,..., x^{\tau}_m\}$ and $\mathrm{\bf x}=\{x_1, x_2,..., x_n\}$, the aim of target sentiment classification is to predict the sentiment polarity $y \in \{P, N, O\}$ of the sentence $\mathrm{\bf w}$ over the target $\mathrm{\bf w}^{\tau}$, where $P$, $N$ and $O$ denote ``positive'', ``negative'' and ``neutral'' sentiments respectively.

The architecture of the proposed Target-Specific Transformation Networks (TNet) is shown in Fig.~\ref{fig:architecture}. The bottom layer is a BiLSTM which transforms the input $\mathrm{\bf x}=\{x_1, x_2,..., x_n\} \in \mathbb{R}^{n \times \mathrm{dim}_w}$ into the contextualized word representations $\mathrm{\bf h}^{(0)}=\{h^{(0)}_1, h^{(0)}_2,..., h^{(0)}_n\} \in \mathbb{R}^{n \times 2\mathrm{dim}_h}$ (i.e. hidden states of BiLSTM), where $\mathrm{dim}_w$ and $\mathrm{dim}_h$ denote the dimensions of the word embeddings and the hidden representations respectively. The middle part, the core part of our TNet, consists of $L$ Context-Preserving Transformation (CPT) layers. The CPT layer incorporates the target information into the word representations via a novel Target-Specific Transformation (TST) component. CPT also contains a context-preserving mechanism, resembling identity mapping~\cite{he2016deep,he2016identity} and highway connection~\cite{srivastava2015training,srivastava2015highway}, allows preserving the context information and learning more abstract word-level features using a deep network. The top most part is a position-aware convolutional layer which first encodes positional relevance between a word and a target, and then extracts informative features for classification.

\subsection{Bi-directional LSTM Layer}
\label{sec:2.2}
\begin{figure}[!t]
    \centering
    \includegraphics[width=0.35\textwidth]{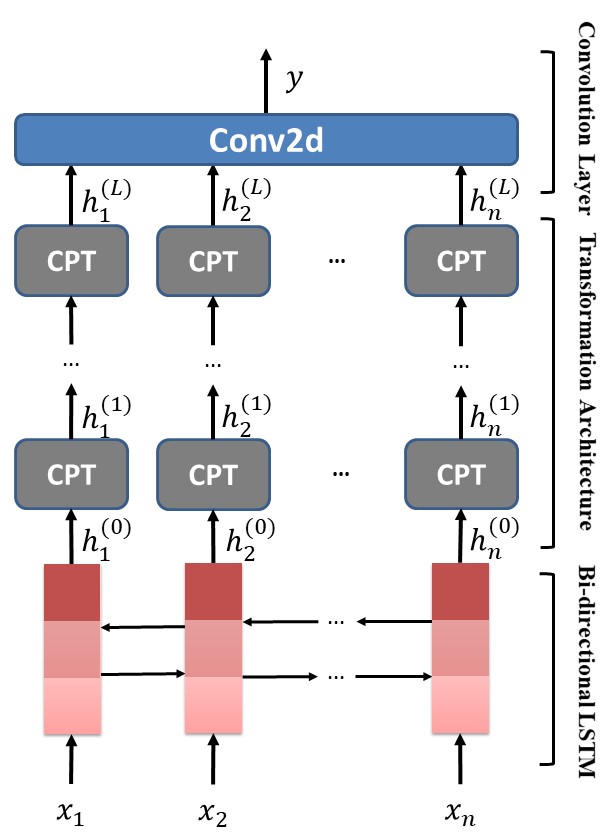}
    \caption{Architecture of TNet.}
    \label{fig:architecture}
\end{figure}

As observed in \citet{lai2015recurrent}, combining contextual information with word embeddings is an effective way to represent a word in convolution-based architectures. 
TNet also employs a BiLSTM to accumulate the context information for each word of the input sentence, i.e., the bottom part in Fig.~\ref{fig:architecture}. 
For simplicity and space issue, we denote the operation of an LSTM unit on $x_i$ as $\text{LSTM}(x_i)$.
Thus, the contextualized word representation $h^{(0)}_i \in \mathbb{R}^{2\mathrm{dim}_h}$ is obtained as follows:
\begin{equation}
h^{(0)}_i = [\overrightarrow{\text{LSTM}}(x_i); \overleftarrow{\text{LSTM}}(x_i)], i \in [1, n].
\end{equation}

\subsection{Context-Preserving Transformation}
\label{sec:cpt}
The above word-level representation has not considered the target information yet. Traditional attention-based approaches keep the word-level features static and aggregate them with weights as the final sentence representation. In contrast, as shown in the middle part in Fig.~\ref{fig:architecture}, we introduce multiple CPT layers and the detail of a single CPT is shown in Fig. \ref{fig:ast}. In each CPT layer, a tailor-made TST component that aims at better consolidating word representation and target representation is proposed. Moreover, we design a context-preserving mechanism enabling the learning of target-specific word representations in a deep neural architecture.

\subsubsection{Target-Specific Transformation}
\begin{figure}[!t]
    \centering
    \includegraphics[width=0.3\textwidth]{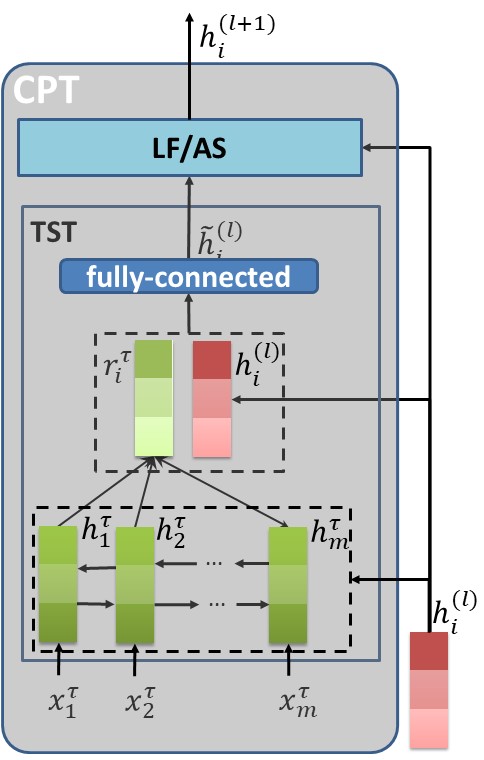}
        \caption{Details of a CPT module.}
    \label{fig:ast}
\end{figure}

TST component is depicted with the TST block in Fig. \ref{fig:ast}.
The first task of TST is to generate the representation of the target. Previous methods \cite{chen-EtAl:2017:EMNLP20171,liu-zhang:2017:EACLshort} average the embeddings of the target words as the target representation. This strategy may be inappropriate in some cases because different target words usually do not contribute equally. For example, in the target ``\textit{amd turin processor}'', the word ``processor'' is more important than ``amd'' and ``turin'', because the sentiment is usually conveyed over the phrase head, i.e.,``processor'', but seldom over modifiers (such as brand name ``amd").
\citet{ma2017interactive} attempted to overcome this issue by measuring the importance score between each target word representation and the averaged sentence vector. 
However, it may be ineffective for sentences expressing multiple sentiments (e.g., \textit{``Air has higher resolution but the fonts are small.''}), because taking the average tends to neutralize different sentiments. 

We propose to dynamically compute the importance of target words based on each sentence word rather than the whole sentence. We first employ another BiLSTM to obtain the target word representations $\mathrm{\bf h}^{\tau} \in \mathbb{R}^{m \times 2\mathrm{dim}_h}$:
\begin{equation}
h^{\tau}_j = [\overrightarrow{\text{LSTM}}(x^{\tau}_j); \overleftarrow{\text{LSTM}}(x^{\tau}_j)], j \in [1, m].
\end{equation}
Then, we dynamically associate them with each word $w_i$ in the sentence to tailor-make target representation $r^{\tau}_i$ at the time step $i$:
\begin{equation}
\label{eq:ast}
\begin{split}
    r^{\tau}_i = \sum^{m}_{j=1} h^{\tau}_j * \mathcal{F}(h^{(l)}_i, h^{\tau}_j)
\end{split},
\end{equation}
where the function $\mathcal{F}$ measures the relatedness between the $j$-th target word representation $h^{\tau}_j$ and the $i$-th word-level representation $h^{(l)}_i$:
\begin{equation}
    \mathcal{F}(h^{(l)}_i, h^{\tau}_j) = \frac{\exp{(h^{(l)\top}_i h^{\tau}_j)}}{\sum^m_{k=1} \exp{(h^{(l)\top}_i h^{\tau}_k})}.
\end{equation}
Finally, the concatenation of $r^{\tau}_i$ and $h^{(l)}_i$ is fed into a fully-connected layer to obtain the $i$-th target-specific word representation $\tilde{h_i}^{(l)}$:
\begin{equation}
\label{eq:transformation}
\tilde{h}^{(l)}_i = g(W^{\tau}[h^{(l)}_i: r^{\tau}_i] + b^{\tau}),
\end{equation}
where $g(*)$ is a non-linear activation function and ``$:$'' denotes vector concatenation. $W^{\tau}$ and $b^{\tau}$ are the weights of the layer.
\subsubsection{Context-Preserving Mechanism}
After the non-linear TST (see Eq.~\ref{eq:transformation}), the context information captured with contextualized representations from the BiLSTM layer
will be lost since the mean and the variance of the features within the feature vector will be changed. To take advantage of the context information, which has been proved to be useful in~\cite{lai2015recurrent}, we investigate two strategies: Lossless Forwarding (LF) and Adaptive Scaling (AS), to pass the context information to each following layer, as depicted by the block ``\textbf{LF/AS}'' in Fig.~\ref{fig:ast}. Accordingly, the model variants are named \textbf{TNet-LF} and \textbf{TNet-AS}.

\paragraph{Lossless Forwarding.}
This strategy preserves context information by directly feeding the features before the transformation to the next layer. Specifically, the input $h^{(l+1)}_i$ of the $(l+1)$-th CPT layer is formulated as:
\begin{equation}
\label{eq:residual1}
    h^{(l+1)}_i = h^{(l)}_i + \tilde{h}^{(l)}_i, i \in [1, n], l \in [0, L],
\end{equation}
where $h^{(l)}_i$ is the input of the $l$-th layer and $\tilde{h}^{(l)}_i$ is the output of TST in this layer. We unfold the recursive form of Eq.~\ref{eq:residual1} as follows:
\begin{equation}
\label{eq:residual2}
    h^{(l+1)}_i = h^{(0)}_i + \text{TST}(h^{(0)}_i) + \cdots + \text{TST}(h^{(l)}_i).
\end{equation}
Here, we denote $\tilde{h}^{(l)}_i$ as $\mathrm{TST}(h^{(l)}_i)$. From Eq.~\ref{eq:residual2}, we can see that the output of each layer will contain the contextualized word representations (i.e., $h^{(0)}_i$), thus, the context information is encoded into the transformed features. We call this strategy ``Lossless Forwarding'' because the contextualized representations and the transformed representations (i.e., $\text{TST}(h^{(l)}_i)$) are kept unchanged during the feature combination.

\paragraph{Adaptive Scaling.}
Lossless Forwarding introduces the context information by directly adding back the contextualized features to the transformed features, which raises a question: Can the weights of the input and the transformed features be adjusted dynamically? With this motivation, we propose another strategy, named ``Adaptive Scaling''. Similar to the gate mechanism in RNN variants~\cite{jozefowicz2015empirical}, Adaptive Scaling introduces a gating function to control the passed proportions of the transformed features and the input features. The gate $\mathrm{\bf t}^{(l)}$ as follows:
\begin{equation}
    t^{(l)}_i = \sigma (W_{trans} h^{(l)}_i + b_{trans}),
\end{equation}
where $t^{(l)}_i$ is the gate for the $i$-th input of the $l$-th CPT layer, and $\sigma$ is the \textit{sigmoid} activation function. Then we perform convex combination of $h^{(l)}_i$ and $\tilde{h}^{(l)}_i$ based on the gate: 
\begin{equation}\label{eq:highway}
    h^{(l+1)}_i = t^{(l)}_i \odot \tilde{h}^{(l)}_i + (1 - t^{(l)}_i) \odot h^{(l)}_i.
\end{equation}
Here, $\odot$ denotes element-wise multiplication. The non-recursive form of this equation is as follows (for clarity, we ignore the subscripts):
\begin{equation}
\label{eq:highway_unfold}
\begin{split}
    &\small{h^{(l+1)} = [\prod^{l}_{k=0} (1-t^{(k)})]\odot h^{(0)}} \\ 
   & \small{ + [t^{(0)}\prod^{l}_{k=1} (1-t^{(k)})]\odot \text{TST}(h^{(0)}) + \cdots} \\
    &\small{+ t^{(l-1)} (1-t^{(l)}) \odot \text{TST}(h^{(l-1)}) 
    + t^{(l)} \odot \text{TST}(h^{(l)}).} \nonumber
\end{split}
\end{equation}

Thus, the context information is integrated in each upper layer and the proportions of the contextualized representations and the transformed representations are controlled by the computed gates in different transformation layers.

\subsection{Convolutional Feature Extractor}
Recall that the second issue that blocks CNN to perform well is that vanilla CNN may associate a target with unrelated general opinion words which are frequently used as modifiers for different targets across domains. For example, ``\textit{\textbf{service}}'' in \textit{``Great food but the service is dreadful''} may be associated with both ``great'' and ``dreadful''. To solve it, we adopt a proximity strategy, which is observed effective in \cite{chen-EtAl:2017:EMNLP20171,li2017deep}. The idea is a closer opinion word is more likely to be the actual modifier of the target.

\begin{table}[t!]
\centering
\resizebox{0.48\textwidth}{!}
{%
\begin{tabular}{ccccc}
\Xhline{3\arrayrulewidth}
 \multicolumn{2}{c}{} & \# Positive & \# Negative & \# Neutral \\ \hline
\multirow{2}{*}{\texttt{LAPTOP}} & Train & 980 & 858 & 454 \\ 
& Test & 340 & 128 & 171 \\ \hline
\multirow{2}{*}{\texttt{REST}} & Train & 2159 & 800 & 632\\ 
& Test & 730 & 195 & 196\\ \hline
\multirow{2}{*}{\texttt{TWITTER}} & Train & 1567 & 1563 & 3127\\ 
& Test & 174 & 174 & 346 \\ \Xhline{3\arrayrulewidth}
\end{tabular}}
\caption{Statistics of datasets.}
\label{tab:statistics}
\end{table}

Specifically, we first calculate the position relevance $v_i$ between the $i$-th word and the target\footnote{As we perform sentence padding, it is possible that the index $i$ is larger than the actual length $n$ of the sentence.}:
\begin{equation}
\label{eq:position}
v_i=
\begin{cases} 
      1-\frac{(k+m-i)}{C} & i < k+m \\
      1-\frac{i-k}{C} & k+m\leq i\leq n \\
      0 & i > n
\end{cases}
\end{equation}
where $k$ is the index of the first target word, $C$ is a pre-specified constant, and $m$ is the length of the target $\mathrm{\bf w}^\tau$. Then, we use $v$ to help CNN locate the correct opinion w.r.t. the given target:
\begin{equation}
\label{eq:multiply}
    \hat{h}^{(l)}_i = h^{(l)}_i * v_i, i \in [1,n], l \in [1,L].
\end{equation}
Based on Eq.~\ref{eq:position} and Eq.~\ref{eq:multiply}, the words close to the target will be highlighted and those far away will be downgraded. $v$ is also applied on the intermediate output to introduce the position information into each CPT layer. Then we feed the weighted $\mathrm{\bf h}^{(L)}$ to the convolutional layer, i.e., the top-most layer in Fig.~\ref{fig:architecture}, to generate the feature map $\mathrm{\bf c} \in \mathbb{R}^{n-s+1}$ as follows:
\begin{equation}
    c_i = \mathrm{ReLU}(\boldsymbol{w}^{\top}_{conv}  \mathrm{\bf h}^{(L)}_{i:i+s-1} + b_{conv}),
\end{equation}
where $\mathrm{\bf h}^{(L)}_{i:i+s-1} \in \mathbb{R}^{s \cdot \mathrm{dim}_h}$ is the concatenated vector of $\hat{h}^{(L)}_i,\cdots,\hat{h}^{(L)}_{i+s-1}$, and $s$ is the kernel size. $\boldsymbol{w}_{conv} \in \mathbb{R}^{s \cdot \mathrm{dim}_h} $ and $b_{conv} \in \mathbb{R}$ are learnable weights of the convolutional kernel. To capture the most informative features, we apply max pooling~\cite{kim:2014:EMNLP2014} and obtain the sentence representation $z \in \mathbb{R}^{n_k}$ by employing $n_k$ kernels:
\begin{equation}
    z = [\mathrm{max}(\mathrm{\bf c}_1), \cdots, \mathrm{max}(\mathrm{\bf c}_{n_k})] ^ {\top}.
\end{equation}
Finally, we pass $z$ to a fully connected layer for sentiment prediction:
\begin{equation}
    p(y|\mathrm{\bf w}^\tau,\mathrm{\bf w}) = \mathrm{Softmax}(W_{f} z + b_{f}).
\end{equation}
where $W_f$ and $b_f$ are learnable parameters.

\begin{table*}[!t]
    \centering
    {%
    \begin{tabular}{l|ccc|ccc}
    \Xhline{3\arrayrulewidth}
    \multirow{2}{*}{\textbf{Hyper-params}} & \multicolumn{3}{c}{\textbf{TNet-LF}} & \multicolumn{3}{|c}{\textbf{TNet-AS}} \\ \cline{2-7}
    & \texttt{LAPTOP} & \texttt{REST} & \texttt{TWITTER} & \texttt{LAPTOP} & \texttt{REST} & \texttt{TWITTER} \\ \hline
    $\mathrm{dim}_w$ & \multicolumn{3}{c}{300} & \multicolumn{3}{|c}{300} \\ 
    $\mathrm{dim}_h$ & \multicolumn{3}{c}{50} & \multicolumn{3}{|c}{50}  \\ 
    dropout rates ($p_{lstm}$, $p_{sent}$) & \multicolumn{3}{|c}{(0.3, 0.3)} & \multicolumn{3}{|c}{(0.3, 0.3)} \\ 
    $L$ & \multicolumn{3}{c}{2} & \multicolumn{3}{|c}{2} \\
    batch size & 64 & 25 & 64 & 64 & 32 & 64 \\ 
    $s$ & \multicolumn{3}{c}{3} & \multicolumn{3}{|c}{3} \\ 
    $n_k$ & \multicolumn{3}{c}{50} & \multicolumn{3}{|c}{100} \\ 
    $C$ & \multicolumn{3}{c}{40.0} & \multicolumn{3}{|c}{30.0} \\
    \Xhline{3\arrayrulewidth}
    \end{tabular}}
    \caption{Settings of hyper-parameters.}
    \label{tab:hyperparameter}
\end{table*}

\section{Experiments}
\subsection{Experimental Setup}
As shown in Table~\ref{tab:statistics}, we evaluate the proposed TNet on three benchmark datasets: \texttt{LAPTOP} and \texttt{REST} are from SemEval ABSA challenge~\cite{pontiki-EtAl:2014:SemEval}, containing user reviews in laptop domain and restaurant domain respectively. We also remove a few examples having the ``conflict label'' as done in~\cite{chen-EtAl:2017:EMNLP20171}; \texttt{TWITTER} is built by \citet{dong-EtAl:2014:P14-2}, containing twitter posts. All tokens are lowercased without removal of stop words, symbols or digits, and sentences are zero-padded to the length of the longest sentence in the dataset. Evaluation metrics are Accuracy and Macro-Averaged F1 where the latter is more appropriate for datasets with unbalanced classes. We also conduct pairwise t-test on both Accuracy and Macro-Averaged F1 to verify if the improvements over the compared models are reliable.


TNet is compared with the following methods.
\begin{itemize}
\item \textbf{SVM}~\cite{kiritchenko-EtAl:2014:SemEval}: It is a traditional support vector machine based model with extensive feature engineering;
\item \textbf{AdaRNN}~\cite{dong-EtAl:2014:P14-2}: It learns the sentence representation toward target for sentiment prediction via semantic composition over dependency tree; 
\item \textbf{AE-LSTM}, and \textbf{ATAE-LSTM}~\cite{wang-EtAl:2016:EMNLP20163}: AE-LSTM is a simple LSTM model incorporating the target embedding as input, while ATAE-LSTM extends AE-LSTM with attention;
\item  \textbf{IAN}~\cite{ma2017interactive}: IAN employs two LSTMs to learn the representations of the context and the target phrase interactively;
\item \textbf{CNN-ASP}: It is a CNN-based model implemented by us which directly concatenates target representation to each word embedding;
\item  \textbf{TD-LSTM}~\cite{tang-EtAl:2016:COLING3}: It employs two LSTMs to model the left and right contexts of the target separately, then performs predictions based on concatenated context representations;
\item  \textbf{MemNet}~\cite{tang-qin-liu:2016:EMNLP2016}: It applies attention mechanism over the word embeddings multiple times and predicts sentiments based on the top-most sentence representations;
\item \textbf{BILSTM-ATT-G}~\cite{liu-zhang:2017:EACLshort}: It models left and right contexts using two attention-based LSTMs and introduces gates to measure the importance of left context, right context, and the entire sentence for the prediction;
\item  \textbf{RAM}~\cite{chen-EtAl:2017:EMNLP20171}: RAM is a multi-layer architecture where each layer consists of attention-based aggregation of word features and a GRU cell to learn the sentence representation. 
\end{itemize}

We run the released codes of TD-LSTM and BILSTM-ATT-G to generate results, since their papers only reported results on \texttt{TWITTER}. We also rerun MemNet on our datasets and evaluate it with both accuracy and Macro-Averaged F1.\footnote{The codes of TD-LSTM/MemNet and BILSTM-ATT-G are available at: \url{http://ir.hit.edu.cn/~dytang} and \url{http://leoncrashcode.github.io}. Note that MemNet was only evaluated with accuracy.}

We use pre-trained GloVe vectors \cite{pennington2014glove} to initialize the word embeddings and the dimension is 300 (i.e., $\mathrm{dim}_w=300$). For out-of-vocabulary words, we randomly sample their embeddings from the uniform distribution $\mathcal{U}(-0.25, 0.25)$, as done in~\cite{kim:2014:EMNLP2014}. 
We only use one convolutional kernel size because it was observed that CNN with single optimal kernel size is comparable with CNN having multiple kernel sizes on small datasets~\cite{zhang-wallace:2017:I17-1}. To alleviate overfitting, we apply dropout on the input word embeddings of the LSTM and the ultimate sentence representation $z$. All weight matrices are initialized with the uniform distribution $\mathcal{U}(-0.01, 0.01)$ and the biases are initialized as zeros. The training objective is cross-entropy, and Adam~\cite{kingma2014adam} is adopted as the optimizer by following the learning rate and the decay rates in the original paper. 

\begin{table*}[!t]

    \centering
    \resizebox{1\textwidth}{!}{%
    \begin{tabular}{llcccccc}
    \Xhline{3\arrayrulewidth}
      & \multirow{2}{*}{\textbf{Models}} & \multicolumn{2}{c}{\texttt{LAPTOP}} & \multicolumn{2}{c}{\texttt{REST}} & \multicolumn{2}{c}{\texttt{TWITTER}}   \\ \cline{3-8}
       & & ACC & Macro-F1 & ACC & Macro-F1 & ACC & Macro-F1 \\ \hline \hline
       \multirow{10}{*}{\textbf{Baselines}} & SVM & 70.49$^\natural$ & - & 80.16$^\natural$ & - & 63.40$^*$ & 63.30$^*$ \\
        & AdaRNN & - & - & - & - & 66.30$^\natural$ & 65.90$^\natural$ \\ 
        &AE-LSTM & 68.90$^\natural$ & - & 76.60$^\natural$ & - & - & - \\
        &ATAE-LSTM & 68.70$^\natural$ & - & 77.20$^\natural$ & - & - & - \\ 
        &IAN & 72.10$^\natural$ & - & 78.60$^\natural$ & - & - & - \\ 
        &CNN-ASP & 72.46 & 65.31 & 77.82 & 65.11 & 73.27 & 71.77 \\
        &TD-LSTM & 71.83 & 68.43 & 78.00 & 66.73 & 66.62 & 64.01 \\
        &MemNet & 70.33 & 64.09 & 78.16 & 65.83 & 68.50 & 66.91  \\
        & BILSTM-ATT-G & 74.37 & 69.90 & 80.38 & 70.78 & 72.70 & 70.84 \\
        & RAM & 75.01 & 70.51 & 79.79 & 68.86 & 71.88 & 70.33 \\\hline 
        \multirow{3}{*}{\textbf{CPT Alternatives}} & LSTM-ATT-CNN & 73.37 & 68.03 & 78.95 & 68.71 & 70.09 & 67.68 \\
        & LSTM-FC-CNN-LF & 75.59 & 70.60 & 80.41 & 70.23 & 73.70 & 72.82 \\ 
        & LSTM-FC-CNN-AS & 75.78 & 70.72 & 80.23 & 70.06 & 74.28 & 72.60 \\
        \hline 
        \multirow{4}{*}{\textbf{Ablated TNet}} & TNet w/o transformation & 73.30 & 68.25 & 78.90 & 65.86 & 72.10 & 70.57 \\
        & TNet w/o context & 73.91 & 68.87 & 80.07 & 69.01 & 74.51 & 73.05 \\ 
        & TNet-LF w/o position & 75.13 & 70.63 & 79.86 & 69.69 & 73.83 & 72.49 \\ 
         & TNet-AS w/o position & 75.27 & 70.03 & 79.79 & 69.78 & 73.84 & 72.47\\  \hline
        \multirow{2}{*}{\textbf{TNet variants}}& TNet-LF & 76.01$^{\dag,\ddag}$ & 71.47$^{\dag,\ddag}$ & \textbf{80.79}$^{\dag,\ddag}$ & 70.84$^{\ddag}$ & 74.68$^{\dag,\ddag}$ & 73.36$^{\dag,\ddag}$ \\ 
        & TNet-AS & \textbf{76.54}$^{\dag,\ddag}$ & \textbf{71.75}$^{\dag,\ddag}$ & 80.69$^{\dag,\ddag}$ & \textbf{71.27}$^{\dag,\ddag}$ & \textbf{74.97}$^{\dag,\ddag}$ & \textbf{73.60}$^{\dag,\ddag}$ \\ 
    \Xhline{3\arrayrulewidth}
    \end{tabular}}
    \caption{Experimental results (\%). The results with symbol``$\natural$'' are retrieved from the original papers, and those starred ($*$) one are from~\citet{dong-EtAl:2014:P14-2}. The marker $^{\dag}$ refers to $p$-value $<$ 0.01 when comparing with BILSTM-ATT-G, while the marker $^{\ddag}$ refers to $p$-value $<$ 0.01 when comparing with RAM. }
    \label{tab:main_results}
\end{table*}

The hyper-parameters of TNet-LF and TNet-AS are listed in Table~\ref{tab:hyperparameter}. Specifically, all hyper-parameters are tuned on 20\% randomly held-out training data 
and the hyper-parameter collection producing the highest accuracy score is used for testing. 
Our model has comparable number of parameters compared to traditional LSTM-based models as we reuse parameters in the transformation layers and BiLSTM.\footnote{All experiments are conducted on a single NVIDIA GTX 1080. The prediction cost of a sentence is about 2 ms.}

\subsection{Main Results}
As shown in Table~\ref{tab:main_results}, both TNet-LF and TNet-AS consistently achieve the best performance on all datasets, which verifies the efficacy of our whole TNet model. Moreover, TNet can perform well for different kinds of user generated content, such as product reviews with relatively formal sentences in \texttt{LAPTOP} and \texttt{REST}, and tweets with more ungrammatical sentences in \texttt{TWITTER}. The reason is the CNN-based feature extractor arms TNet with more power to extract accurate features from ungrammatical sentences. Indeed, we can also observe that another CNN-based baseline, i.e., CNN-ASP implemented by us, also obtains good results on \texttt{TWITTER}. 

On the other hand, the performance of those comparison methods is mostly unstable. For the tweet in \texttt{TWITTER}, the competitive BILSTM-ATT-G and RAM cannot perform as effective as they do for the reviews in \texttt{LAPTOP} and \texttt{REST}, due to the fact that they are heavily rooted in LSTMs and the ungrammatical sentences hinder their capability in capturing the context features. Another difficulty caused by the ungrammatical sentences is that the dependency parsing might be error-prone, which will affect those methods such as AdaRNN using dependency information.

From the above observations and analysis, some takeaway message for the task of target sentiment classification could be: 
\begin{itemize}
\item LSTM-based models relying on sequential information can perform well for formal sentences by capturing more useful context features; 
\item For ungrammatical text, CNN-based models may have some advantages because CNN aims to extract the most informative n-gram features and is thus less sensitive to informal texts without strong sequential patterns.
\end{itemize}



\subsection{Performance of Ablated TNet}


To investigate the impact of each component such as deep transformation, context-preserving mechanism, and positional relevance, we perform comparison between the full TNet models and its ablations (the third group in Table~\ref{tab:main_results}). After removing the deep transformation (i.e., the techniques introduced in Section~\ref{sec:cpt}), both TNet-LF and TNet-AS are reduced to TNet w/o transformation (where position relevance is kept), and their results in both accuracy and F1 measure are incomparable with those of TNet. It shows that the integration of target information into the word-level representations is crucial for good performance. 

Comparing the results of TNet and TNet w/o context (where TST and position relevance are kept), we observe that the performance of TNet w/o context drops significantly on \texttt{LAPTOP} and \texttt{REST}\footnote{Without specification, the significance level is set to 0.05.}, while on \texttt{TWITTER}, TNet w/o context performs very competitive ($p$-values with TNet-LF and TNet-AS are 0.066 and 0.053 respectively for Accuracy). Again, we could attribute this phenomenon to the ungrammatical user generated content of twitter, because the context-preserving component becomes less important for such data. TNet w/o context performs consistently better than TNet w/o transformation, which verifies the efficacy of the target specific transformation (TST), before applying context-preserving. 

As for the position information, we conduct statistical t-test between TNet-LF/AS and TNet-LF/AS w/o position together with performance comparison. All of the produced $p$-values are less than 0.05, suggesting that the improvements brought in by position information are significant. 


\subsection{CPT versus Alternatives}
The next interesting question is what if we replace the transformation module (i.e., the CPT layers in Fig.\ref{fig:architecture}) of TNet with other commonly-used components? We investigate two alternatives: attention mechanism and fully-connected (FC) layer, resulting in three pipelines as shown in the second group of Table~\ref{tab:main_results} (position relevance is kept for them).

LSTM-ATT-CNN applies attention as the alternative\footnote{We tried different attention mechanisms and report the best one here, namely, \textit{dot} attention~\cite{luong-pham-manning:2015:EMNLP}.}, and it does not need the context-preserving mechanism. It performs unexceptionally worse than the TNet variants. We are surprised that LSTM-ATT-CNN is even worse than TNet w/o transformation (a pipeline simply removing the transformation module) on \texttt{TWITTER}. More concretely, applying attention results in negative effect on \texttt{TWITTER}, which is consistent with the observation that all those attention-based state-of-the-art methods (i.e., TD-LSTM, MemNet, BILSTM-ATT-G, and RAM) cannot perform well on \texttt{TWITTER}. 

LSTM-FC-CNN-LF and LSTM-FC-CNN-AS are built by applying FC layer to replace TST and keeping the context-preserving mechanism (i.e., LF and AS). Specifically, the concatenation of word representation and the averaged target vector is fed to the FC layer to obtain target-specific features. Note that LSTM-FC-CNN-LF/AS are equivalent to TNet-LF/AS when processing single-word targets (see Eq.~\ref{eq:ast}).
They obtain competitive results on all datasets: comparable with or better than the state-of-the-art methods. The TNet variants can still outperform LSTM-FC-CNN-LF/AS with significant gaps, e.g., on \texttt{LAPTOP} and \texttt{REST}, the accuracy gaps between TNet-LF and LSTM-FC-CNN-LF are 0.42\% ($p <$ 0.03) and 0.38\% ($p <$ 0.04) respectively. 

\begin{figure}[!t]
    \centering
    \includegraphics[width=0.5\textwidth]{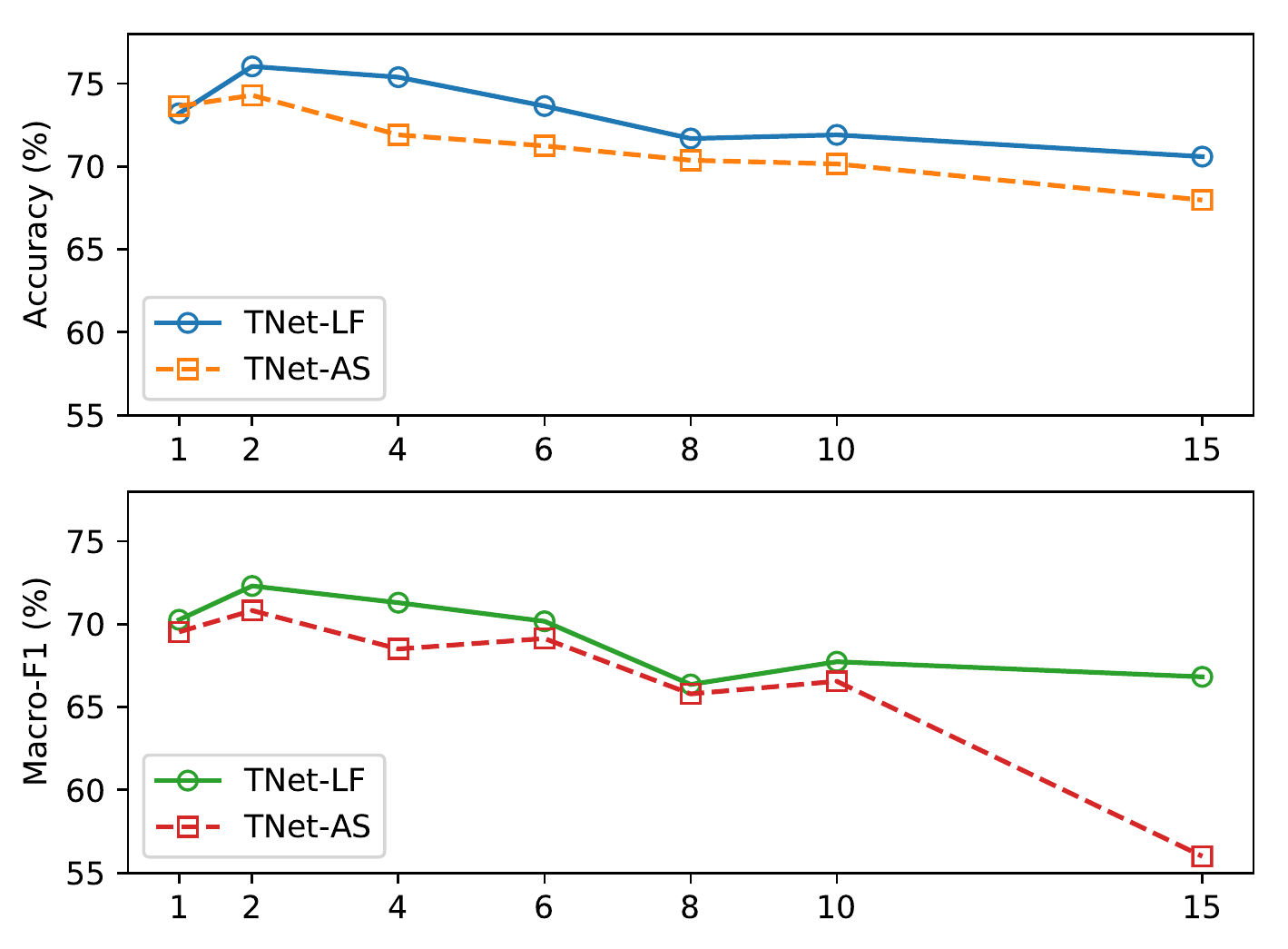}
    \caption{Effect of $L$.}
    \label{fig:layer}
\end{figure}

\subsection{Impact of CPT Layer Number}
As our TNet involves multiple CPT layers, we investigate the effect of the layer number $L$. Specifically, we conduct experiments on the held-out training data of \texttt{LAPTOP} and vary $L$ from 2 to 10, increased by 2. The cases $L$=1 and $L$=15 are also included. The results are illustrated in Figure~\ref{fig:layer}. We can see that both TNet-LF and TNet-AS achieve the best results when $L$=2. While increasing $L$, the performance is basically becoming worse. For large $L$, the performance of TNet-AS generally becomes more sensitive, it is probably because AS involves extra parameters (see Eq~\ref{eq:highway}) that increase the training difficulty.  

\begin{table*}[!t]
    \centering
    \resizebox{0.99\textwidth}{!}
    {%
    \begin{tabular}{L{9.5cm}||cccc}
   \Xhline{3\arrayrulewidth}
        \textbf{Sentence} & \textbf{BILSTM-ATT-G} & \textbf{RAM} & \textbf{TNet-LF} & \textbf{TNet-AS} \\ \hline
        1. \textcolor{orange}{Air has higher} \setulcolor{orange}\ul{\textbf{[resolution]}}$_\text{P}$ but the \textcolor{blue}{\setulcolor{blue}\ul{\textbf{[fonts]}}$_\text{N}$ are small} . & (N$^{\text{\xmark}}$, N) & (N$^{\text{\xmark}}$, N) & (P, N) & (P, N) \\ \hline
        2. \textcolor{orange}{Great \setulcolor{orange}\ul{\textbf{[food]}}$_\text{P}$ but} the \setulcolor{blue}\ul{\textbf{[service]}}$_\text{N}$ is \textcolor{blue}{dreadful .} & (P, N) & (P, N) & (P, N) & (P, N) \\ \hline
        3. Sure it ' s not light and slim but the \textcolor{orange}{\setulcolor{orange}\ul{\textbf{[features]}}$_\text{P}$ make up} for it 100\% . & N$^{\text{\xmark}}$ & N$^{\text{\xmark}}$ & P & P \\ \hline
        4. Not only did they have \textcolor{orange}{amazing , \setulcolor{orange}\ul{\textbf{[sandwiches]}}$_\text{P}$} , \setulcolor{orange}\ul{\textbf{[soup]}}$_\text{P}$ , \setulcolor{orange}\ul{\textbf{[pizza]}}$_\text{P}$ etc , but their \setulcolor{blue}\ul{\textbf{[homemade sorbets]}}$_\text{P}$ are out of \textcolor{blue}{this world !} & (P, O$^{\text{\xmark}}$, O$^{\text{\xmark}}$, P) & (P, P, O$^{\text{\xmark}}$, P) &  (P, P, P, P) & (P, P, P, P) \\ \hline
        5. \setulcolor{orange}\ul{\textbf{[startup times]}}$_\text{N}$ are \textcolor{orange}{incredibly long :} over two minutes . & P$^{\text{\xmark}}$ & P$^{\text{\xmark}}$ &  N & N \\ \hline
        6. I am pleased with the \textcolor{orange}{fast \setulcolor{orange}\ul{\textbf{[log on]}}$_\text{P}$} , \textcolor{blue}{speedy \setulcolor{blue}\ul{\textbf{[wifi connection]}}$_\text{P}$} and the \textcolor{green}{long \setulcolor{green}\ul{\textbf{[battery life]}}$_\text{P}$} ( $>$ 6 hrs ) . & (P, P, P) & (P, P, P) & (P, P, P) & (P, P, P) \\ \hline
        7. The \setulcolor{orange}\ul{\textbf{[staff]}}$_\text{N}$ should be a bit more \textcolor{orange}{friendly .} & P$^{\text{\xmark}}$ & P$^{\text{\xmark}}$ & P$^{\text{\xmark}}$ & P$^{\text{\xmark}}$ \\ \hline 
    \Xhline{3\arrayrulewidth}
    \end{tabular}}
        \caption{Example predictions, color printing is preferred. The input targets are wrapped in brackets with the true labels given as subscripts. ${\text{\xmark}}$ indicates incorrect prediction.}
    \label{tab:case_study}
\end{table*}

\subsection{Case Study}
Table~\ref{tab:case_study} shows some sample cases. 
The input targets are wrapped in the brackets with true labels given as subscripts. The notations P, N and O in the table represent positive, negative and neutral respectively. For each sentence, we underline the target with a particular color, and the text of its corresponding most informative n-gram feature\footnote{For each convolutional filter, only one n-gram feature in the feature map will be kept after the max pooling. Among those from different filters, the n-gram with the highest frequency will be regarded as the most informative n-gram w.r.t. the given target.} captured by TNet-AS (TNet-LF captures very similar features) is in the same color (so color printing is preferred). For example, for the target ``resolution'' in the first sentence, the captured feature is ``Air has higher''. 
Note that as discussed above, the CNN layer of TNet captures such features with the size-three kernels, so that the features are trigrams. Each of the last features of the second and seventh sentences contains a padding token, which is not shown. 

Our TNet variants can predict target sentiment more accurately than RAM and BILSTM-ATT-G in the transitional sentences such as the first sentence by capturing correct trigram features. For the third sentence, its second and third most informative trigrams are ``100\% . PAD'' and ``' s not'', being used together with ``features make up'', our models can make correct predictions. 
Moreover, TNet can still make correct prediction when the explicit opinion is target-specific. For example, ``long'' in the fifth sentence is negative for ``startup time'', while it could be positive for other targets such as ``battery life'' in the sixth sentence. The sentiment of target-specific opinion word is conditioned on the given target. Our TNet variants, armed with the word-level feature transformation w.r.t. the target, is capable of handling such case.

We also find that all these models cannot give correct prediction for the last sentence, a commonly used subjunctive style. In this case, the difficulty of prediction does not come from the detection of explicit opinion words but the inference based on implicit semantics, which is still quite challenging for neural network models.




\section{Related Work}

Apart from sentence level sentiment classification \cite{kim:2014:EMNLP2014,bei2018sentiment_embedding}, aspect/target level sentiment classification 
is also an important research topic in the field of sentiment analysis. The early methods mostly adopted supervised learning approach with extensive hand-coded features~\cite{blair2008building,titov2008modeling,yu2011aspect,jiang-EtAl:2011:ACL-HLT2011,kiritchenko-EtAl:2014:SemEval,wagner-EtAl:2014:SemEval,vo2015target}, and they fail to model the semantic relatedness between a target and its context which is critical for target sentiment analysis. \citet{dong-EtAl:2014:P14-2} incorporate the target information into the feature learning using dependency trees. As observed in previous works, the performance heavily relies on the quality of dependency parsing. \citet{tang-EtAl:2016:COLING3} propose to split the context into two parts and associate target with contextual features separately. Similar to \cite{tang-EtAl:2016:COLING3}, \citet{zhang2016gated} develop a three-way gated neural network to model the interaction between the target and its surrounding contexts. Despite the advantages of jointly modeling target and context, they are not capable of capturing long-range information when some critical context information is far from the target. To overcome this limitation, researchers bring in the attention mechanism to model target-context association~\cite{tang-EtAl:2016:COLING3,tang-qin-liu:2016:EMNLP2016,wang-EtAl:2016:EMNLP20163,yang2017attention,liu-zhang:2017:EACLshort,ma2017interactive,chen-EtAl:2017:EMNLP20171,zhang-EtAl:2017:I17-11,tay2017learning}. 
Compared with these methods, our TNet avoids using attention for feature extraction so as to alleviate the attended noise. 

\section{Conclusions}
We re-examine the drawbacks of attention mechanism for target sentiment classification, and also investigate the obstacles that hinder CNN-based models to perform well for this task. Our TNet model is carefully designed to solve these issues. Specifically, we propose target specific transformation component to better integrate target information into the word representation. Moreover, we employ CNN as the feature extractor for this classification problem, and rely on the context-preserving and position relevance mechanisms to maintain the advantages of previous LSTM-based models. The performance of TNet consistently dominates previous state-of-the-art methods on different types of data. 
The ablation studies show the efficacy of its different modules, and thus verify the rationality of TNet's architecture.

\bibliography{acl2018}
\bibliographystyle{acl_natbib}

\end{document}